\documentclass[letterpaper,10pt,conference]{ieeeconf}

\IEEEoverridecommandlockouts
\let\labelindent\relax
\usepackage{enumitem}
\usepackage{amssymb}
\usepackage{xcolor}
\usepackage{booktabs}
\usepackage{multirow}
\usepackage{float}
\usepackage{lipsum}
\usepackage{graphicx}
\usepackage{amsmath}
\usepackage[mathcal]{eucal}
\usepackage{caption}
\usepackage{stfloats}
\usepackage{cuted}
\usepackage{cite}
\usepackage{xspace}
\usepackage{indentfirst}
\usepackage{arydshln}
\usepackage[pdftex,hidelinks]{hyperref}
\hypersetup{
	colorlinks=false,
	pdfborder={0 0 0},
	breaklinks=true
}
\usepackage{orcidlink}
\usepackage{fancyhdr}
\usepackage{eso-pic}

\makeatletter

\let\ACorigbibitem\bibitem
\def\bibitem{\@ifnextchar[\AClbibitem\ACbibitem}

\def\ACbibitem#1{%
	\ACorigbibitem{#1}%
	\hypertarget{bib:#1}{}%
}

\def\AClbibitem[#1]#2{%
	\ACorigbibitem[#1]{#2}%
	\hypertarget{bib:#2}{}%
}

\makeatother

\newcommand{\linkcite}[1]{\unskip\hyperlink{bib:#1}{\cite{#1}}\xspace}

\definecolor{boldcolor}{gray}{0.2} % range from [0,1]

\begin{document}

\title{\LARGE \bf
Multimodal Voice Activity Projection for Social Robot Mediation: Expected Behavior and Deployment Constraints
%From Multimodal Voice Activity Projection to Social Robot Mediation Behavior
}

\author{Antonio Cano$^{1,2}$\,\orcidlink{0000-0002-0435-4987}\label{author:AC},
	Guillermo Pérez$^{1}$\,\orcidlink{0000-0002-8358-996X}\label{author:GP},
	Luis Merino$^{2}$\,\orcidlink{0000-0003-4927-8647}\label{author:LM} and
	Randy Gomez$^{3}$\,\orcidlink{0000-0002-3191-6818}\label{author:RG}
	\thanks{$^{1}$ Antonio Cano and Guillermo Perez are with 4i Intelligent Insights, Seville, Spain.
    {\tt\small \href{mailto:a.cano@4i.ai}{a.cano@4i.ai},
    \href{mailto:g.perez@4i.ai}{g.perez@4i.ai}}}%
    \thanks{$^{3}$ Antonio Cano is with Universidad de Sevilla, Seville, Spain.
    {\tt\small \href{mailto:acano4@us.es}{acano4@us.es}}}%
    \thanks{$^{2}$ Luis Merino is with Universidad Pablo de Olavide, Seville, Spain.
    {\tt\small \href{mailto:lmercab@upo.es}{lmercab@upo.es}}}%
    \thanks{$^{3}$ Randy Gomez is with Honda Research Institute Japan, Saitama, Japan.
    {\tt\small \href{mailto:r.gomez@jp.honda-ri.com}{r.gomez@jp.honda-ri.com}}}%
}

\maketitle

\pagestyle{fancy}
\fancyhf{}

% Header
\AddToShipoutPictureFG*{%
	\AtPageUpperLeft{%
		\raisebox{-1.5cm}[0pt][0pt]{%
			\makebox[\paperwidth][c]{%
				\begin{minipage}{\paperwidth}
					\centering
					\normalsize
					Presented at IEEE RO-MAN 2026 at\\
					3rd Workshop on Nonverbal Cues for Human-Robot Cooperative Intelligence (NOC)\\
					\textbf{Best Workshop Paper Award}
				\end{minipage}%
			}%
		}%
	}%
}

% Footer
\fancyfoot[C]{\normalsize
	3rd Workshop on Nonverbal Cues for Human-Robot Cooperative Intelligence (NOC)}
\renewcommand{\headrulewidth}{0pt}
\renewcommand{\footrulewidth}{0pt}

\begin{abstract}
%\textcolor{red}{\lipsum[1]}

%Turn-taking prediction is a key requirement for social robots involved in human-human interaction, particularly when the robot acts as a mediator rather than as the main interlocutor. This paper reframes Multimodal Voice Activity Projection (MM-VAP) as a human state-aware perception module for social robot mediation. The model estimates future conversational floor dynamics from synchronized audio-visual streams, preserving the self-supervised projection objective of Voice Activity Projection while incorporating voice-activity-related pretrained audio-visual encoders. The proposed formulation uses Low-Rank Adaptation to specialize the backbones and inter-speaker attention to model the relational dynamics between interlocutors. In addition to the predictive events evaluated in the original framework, \color{red}{this version} defines an expected interface between predicted turn-taking events and downstream robot mediation outputs, including gaze control, overlap interpretation, and balanced intervention strategies. The approach is evaluated on multilingual dyadic interaction data and on the Haru EDR corpus. The main deployment limitation remains the real-time integration cost of large multimodal encoders, where inference latency, memory consumption, and end-to-end response time must be measured in the final robotic setup.

Turn-taking prediction is especially relevant for social robots that act as mediators in human-human interaction, where the expected action is often not to speak, but to orient, wait, avoid interruption, or prepare a balanced intervention. This paper presents Multimodal Voice Activity Projection (MM-VAP) as a human state-aware perception layer for future robot mediation behavior. The model estimates the future evolution of the conversational floor from synchronized audio-visual evidence and derives turn-taking events such as Hold, Shift, Shift prediction, Backchannel prediction, and overlap-related states. The approach uses VA-related pretrained audio-visual encoders, LoRA adaptation, inter-speaker attention, and zero-shot event inference from future voice activity projections. Experiments on NoXi, NoXi+J, and Haru EDR support the feasibility of this formulation, especially for floor management events that can be connected to gaze preparation, active listening, and conservative intervention. Finally, the paper defines the expected robot output interface and discusses the main deployment constraints, including real-time inference, preprocessing latency, multimodal synchronization, and input-quality monitoring.

\vspace{-0.5mm}

\noindent The source codes and
pretrained models are available at \textcolor{gray}{\href{https://github.com/acano15/MM-VAP}{https://github.com/acano15/MM-VAP}}.
\end{abstract}
%\vspace{-3.8mm}
%\vspace{-0.5mm}
\section{Introduction}
\vspace{-1mm}
Turn-taking is the conversational mechanism that structures spoken interaction and coordinates the roles of speaker and listener \linkcite{sacksSimplestSystematicsOrganization1974}. In social robotics, this mechanism is especially relevant because the robot must not only recognize the end of an utterance, but also interpret the evolution of the conversational floor over time. This requirement becomes more demanding when the robot is placed in a human-human interaction as a background mediator, where an incorrect intervention may interrupt a speaker, reinforce an overlap, or reduce the naturalness of the dialogue.
Most practical dialogue systems still rely on reactive end-of-turn decisions \linkcite{skantzeTurntakingConversationalSystems2021} based on silence or voice activity thresholds. These strategies are useful as low-level safeguards, but they are limited for interaction management because human speakers may pause inside their own turn, produce short backchannels, overlap with another interlocutor, or indicate a transition before silence is actually observed \linkcite{meyerTimingConversation2023}. In addition, it is well established that speech production takes longer than typical turn-transition gaps, which means that humans rely on prediction to prepare their responses before the current speaker’s turn is complete.
Human turn-taking is therefore better described as a predictive process in which participants continuously estimate whether the current floor will be held, shifted, closed, or briefly supported by listener feedback.

The present work builds on the MM-VAP formulation proposed for the Haru social robot mediation scenario \cite{cruzWhenHowExpress2025,cooperDesignSocialFeatures2024}. Haru is not conceived here as one of the main interlocutors, but as an embodied social mediator that observes a dyadic exchange and may support the interaction by regulating conversational flow, balancing participation, managing silences, and expressing socio-emotional behaviors such as active listening. This motivates a representation of turn-taking events not only as benchmark labels, but also as human state estimates that can be consumed by a future robot behavior layer.
This work focus on the applications of a MM-VAP model in the context of robot mediation. First, it summarizes the MM-VAP architecture as a compact audio-visual predictive model for conversational floor estimation. Second, it defines an explicit interface between predicted turn-taking events and expected mediation outputs, with special emphasis on gaze and intervention timing. Third, it discusses the real-time deployment constraints that must be solved before the model can be used as an online module in a physical social robot.

\section{Related Work} \label{related}
Turn-taking prediction has been extensively studied in spoken dialogue systems and human-robot interaction \linkcite{skantzeTurntakingConversationalSystems2021}. Classical approaches usually detect the end of speech from acoustic activity and silence duration, but such decisions are inherently delayed and sensitive to pauses, hesitations, and overlapping speech. Predictive turn-taking models (PTTMs) \linkcite{pintoPredictiveTurnTakingLeveraging2024} address this limitation by estimating future voice activity at frame level, allowing the system to prepare a response before the end of the current turn is fully observed.
Voice Activity Projection (VAP) \linkcite{ekstedtVoiceActivityProjection2022} formulated turn-taking prediction as a self-supervised task that projects the future binary voice activity (VA) of two speakers over a short temporal horizon. The original audio-based model combines a pretrained CPC encoder on raw waveforms with a voice-activity history encoder and a transformer that predicts future activity states. Downstream events such as shift, hold, backchannel, and end-of-turn prediction are derived by aggregating probability mass over interpretable future activity patterns. This formulation been extended to multi-party interaction, backchannel prediction, multilingual settings, and real-time implementations. Recent multimodal extensions of VAP \linkcite{onishiMultimodalVoiceActivity2023,onishiMultimodalVoiceActivity2025,sagaVoiceActivityProjection2025,russellVisualCuesEnhance2025} have shown that non-verbal information can improve the prediction of conversational events. Previous approaches introduced gaze, facial features, head pose, action units, or generic visual encoders as additional sources of evidence. However, most approaches still rely on engineered descriptors or visual backbones not explicitly pretrained for voice-activity-related dynamics, which limits their suitability for modeling audio-visual turn-taking dependencies. In addition, when audio and visual streams are encoded independently, the model must learn cross-modal relations only during VAP training, often from relatively limited conversational data.

%Parameter-efficient adaptation is also relevant in this setting, where large pretrained audio-visual encoders \cite{taoSomeoneSpeakingExploring2021, rouditchenkoWhisperFlamingoIntegratingVisual2024} can be computationally expensive to fine-tune and deploy, but their representations already contain useful speech-related structure. Recent multimodal speech and language research, for instance \linkcite{maScenesMechanisticInterpretability2026,chenGRPOGuidedModalitySelection}, has shown that large pretrained models can be efficiently adapted with Low-Rank Adaptation (LoRA), allowing transfer knowledge to VAP tasks by keeping the original weights frozen and training only a reduced set of low-rank parameters.

%\vspace{-1mm}
\section{Proposed Method: Multimodal Voice Activity Projection (MM-VAP)}
%\vspace{-1mm}
\begin{figure*}[tbh!]
	\centering
	\noindent
	\includegraphics[width=\textwidth,height=60mm]{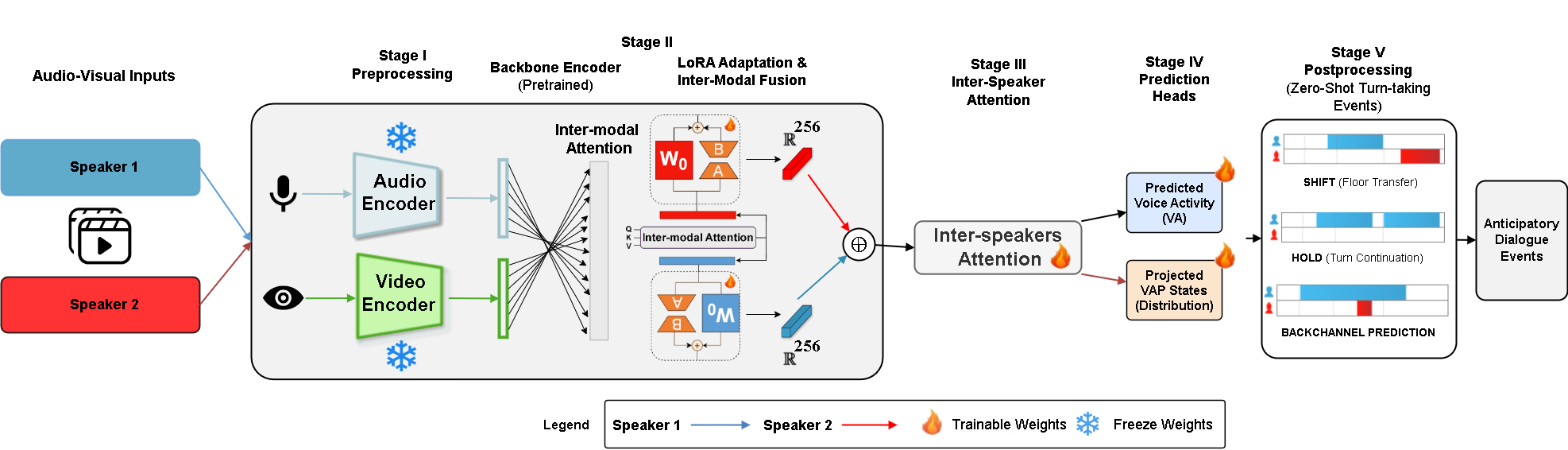}
	%\captionsetup{skip=0pt}
	\caption{Proposed MM-VAP architecture for dyadic interaction overview. Blue and red paths indicate the flow from raw inputs to latent representations for Speaker 1 and Speaker 2.}
	\label{fig:mmvap_pipelilne}
	\vspace*{-5.85mm}
\end{figure*}

This section describes the proposed MM-VAP framework and its complete pipeline, from synchronized audio-visual input sequences to VA and VAP state projection estimation and final inference of turn-taking events.

VAP was formally defined as the task of predicting, or projecting, the future binary VA state of each interlocutor in a dialogue, where each speaker is represented as either active or inactive. MM-VAP extends this formulation by incorporating synchronized audio and visual signals for each interlocutor. The projection is defined over a finite temporal horizon \(\Delta t\), spanning from the current frame \(t\) to a future time \(t+\Delta t\). Within this interval, the future binary VA of both speakers is modeled jointly as a single projection window, \(w_t^{\Delta t}\), which represents the expected future activity trajectory of the dialogue. The length of this projection window must be selected to capture relevant turn-taking dynamics without representing excessively long or complex future interaction patterns. Following the original VAP implementation, \(w_t^{\Delta t}\) is divided into four heterogeneous temporal bins per speaker. The first bins cover the immediate and near future, whereas the later bins represent longer-term regions of the future activity trajectory. Since the interaction involves two speakers, the combination of the four bins for both interlocutors defines a joint output space of \(2^{2N}=256\) possible states. Each state compactly represents one possible future configuration of the interaction horizon between the two interlocutors.

As shown in the Fig. \ref{fig:mmvap_pipelilne} the pipeline is composed of preprocessing, backbone encoding, inter-speaker attention mechanism, prediction heads and turn-taking event zero shot inference. First, the raw audio-visual streams are segmented into speaker-specific context window centered and sincronized at each frame step. For each speaker, the audio chunk waveform is transformed into the spectral representation and the faces sequences are extracted by the face detection module. Both modalities are then synchronized to the same temporal resolution and prepare for the input's requirement of the selected backbone. The backbone selection was based on two criteria: first, the backbone should have achieved high accuracy on its original task related to VA, and second, it must include a robust internal attention mechanism between audio-visual modalities. Therefore, inside the encoder, each modality is processed by its corresponding modality branch, yielding speaker representations in their respective latent spaces. The resulting audio and visual embeddings are fused with the backbone's original inter-modality attention, producing one multimodal embedding per speaker.

In this work, TalkNet and WhisperFlamingo were considered suitable backbone candidates because they satisfy both requirements and provide well-proved pretrained multimodal encoders for ASD and AVSR, respectively, which are closely enough to VA and audio-visual speech understanding. Nevertheless, the original pretrained weights remain specialized for their original task and fully fine-tuning the complete backbone is computationally expensive and may also modify useful pretrained representations. For this reason, LoRA is applied to redirect the capabilities acquired during the original training phase toward future VA projection. LoRA keeps the original weights frozen and introduces trainable low-rank matrices in selected layers reducing the number of training parameters while preserving the original structure and capabilities of the backbone.

Once each speaker embedding is obtained, the model applies a second multihead cross-attention mechanism between both speakers, referred to as inter-speaker attention, to learn how the current multimodal state of one participant influences the other. This stage allows the model to encode turn-taking as a relational process, so the prediction
is conditioned on both the speaker’s current state and the other speaker’s state. The resulting joint representation is then fed to two prediction heads. The first head estimates the multiclass VAP state distribution over the 256 future activity states, while the second head predicts the binary VA probabilities for each speaker. These two outputs are complementary: the VA head provides a direct activity estimation, whereas the VAP head preserves the future-projection structure needed to infer turn-taking events. Finally, the predicted VAP distribution is postprocessed to obtain the target turn-taking event categories in a zero-shot manner by aggregating the probability mass of the states associated with each conversational outcome. This inference follows the original self-supervised VAP strategy, where the model learns from future VA patterns derived directly from the dialogue signal and derives higher-level turn-taking events from the predicted VAP state distribution. Shift and Hold describe whether the conversational floor is expected to move to the other speaker or remain with the current one. Shift prediction estimates whether the other speaker is likely to take the floor soon, even while the current speaker is still active. Backchannel prediction estimates whether the listener is likely to produce a short feedback response without taking the floor. Overlap-related states indicate that both speakers may become active within the prediction horizon.

The loss combines VA estimation, future VAP state projection, and semantic regularization of the 256-state space:
\begin{equation}
    \mathcal{L} = \mathcal{L}_{VA} + \mathcal{L}_{VAP} + \mu \mathcal{L}_{sem}.
    \label{eq:mmvap_loss}
\end{equation}
%
%Here, \(\mathcal{L}_{VA}\) is the multi-label binary loss for VA prediction, \(\mathcal{L}_{VAP}\) is the cross-entropy loss over the 256 future activity states, and \(\mathcal{L}_{sem}\) encourages probability mass to be assigned to states with the same higher-level speaker-activity meaning. This reduces the effect of rare or redundant microstates when deriving turn-taking events.
Here, \(\mathcal{L}_{VA}\) is the binary VA loss, \(\mathcal{L}_{VAP}\) is the cross-entropy loss over the 256 future states, and \(\mathcal{L}_{sem}\) groups semantically equivalent states, reducing redundant microstates influence.
\section{MM-VAP for Haru Mediation: Expected Outputs and Deployment Constraints}

In a social robot mediation scenario, the robot requires an intermediate module that connects conversational floor estimation with observable robot behavior. MM-VAP can provide this connection by producing event-level predictions about the current and future state of the interaction. These predictions can then be translated by the robot controller into gaze, waiting, or mediation actions according to the dialogue context. In this formulation, MM-VAP is treated as a predictive perception layer that estimates who is speaking, who is likely to speak next, and whether the conversational floor is stable, changing, overlapping, or supported by listener feedback.

The most direct output is active-speaker-oriented gaze. When one participant is detected as the active speaker and the model predicts that the floor will be held, the corresponding robot response should be maintain gaze toward that participant, supporting active listening and keeping the robot visually aligned with the current speaker. If a Shift is estimated, the same interface can activate the preparation of a gaze transition toward the expected next speaker. Overlapping speech is associated with a different output: when both speakers become active and the predicted state suggests an interruption-like transition, the proposed gaze target becomes the participant who is taking or contesting the floor. In contrast, backchannel predictions are interpreted as listener feedback, so the current floor configuration is preserved and intervention is suppressed. Together, these gaze updates make the robot’s attention consistent with the participant who becomes interactionally relevant and contribute to the perception of the robot as a real engaged mediator.

When the robot has an opportunity to mediate, the expected output should depend on the dialogue history. After sustained silence, repeated floor dominance, or clear participation imbalance, the robot work is to orient toward the less talkative participant before prompting or yielding the floor. When both participants have spoken with similar balance, the robot can alternate gaze between both interlocutors or address the group without privileging one speaker. Therefore, the MM-VAP event stream would get benefits from the combination with longer-term interaction variables such as accumulated speaking time, number of floor acquisitions, silence duration, and overlap frequency. The perception module estimates the conversational floor dynamics, and the robot controller decides whether the robot should maintain gaze, prepare a gaze transition, wait, suppress intervention, or initiate a mediation turn.

The deployment protocol should also indicate the quality of the input signals. Missing faces, unstable tracking, robot self-noise, room noise, occlusions, or speakers outside the camera view can reduce the reliability of the prediction. For this reason, the robot controller should receive information about which modalities are available, together with the predicted event. When the input is incomplete or degraded, the behavior controller should apply more conservative mediation rules.

Real-time deployment is the most challenging aspect of this proposal. To date, no previous work has demonstrated a similar multimodal turn-taking projection system operating online in a physical robot scenario. In a real Haru setting, the system must deal with the natural behavior of users, which makes the interaction less controlled than in offline experiments. At the same time, several demanding modules must run in parallel, including video acquisition, face tracking, audio processing, modality synchronization, event filtering, and robot control. This creates computational and hardware constraints that may compromise inference time, preprocessing latency, memory usage, synchronization, and robustness. Therefore, future work must evaluate the complete online interaction pipeline and assess its real applicability.

\section{Experiments and results} \label{results}
The experimental protocol follows the original VAP evaluation. The model is trained on synchronized audio-visual context windows and predicts a 2 s future VA. The main evaluation metrics are accuracy and F1-score for event categories associated with floor development, such as shift/hold, short/long responses, shift prediction, and backchannel prediction.
NoXi \linkcite{cafaroNoXiDatabaseMultimodal2017} and NoXi+J \linkcite{funkMultilingualDyadicInteraction2024a} provide multilingual dyadic interaction data with expert-novice conversations. These corpora contain natural two-party interaction and synchronized multimodal recordings. The Haru EDR corpus \linkcite{cruzWhenHowExpress2025} provides the robot-mediated scenario of interest, with semi-structured speaker-listener exchanges conducted in the presence of Haru.

The results indicate that VA-related pretrained backbones provide a clear advantage over the CPC+3DResNet baseline reference. On NoXi, WhisperFlamingo achieved F1 values of 0.97 for both S/H and S-pred, while TalkNet obtained 0.94 for S/H and 0.93 for S/L. On the complete NoXi+J setting, TalkNet provided the strongest BC-pred score among the reported configurations, whereas WhisperFlamingo achieved the strongest S-pred score. These patterns support the hypothesis that pretrained audio-visual representations associated with speech activity can be effectively transferred to future conversational floor projection.

\begin{table}[!h]
\centering
\resizebox{\columnwidth}{!}{%
\begin{tabular}{llcccc}
\hline
\textbf{Corpus} & \textbf{Backbone} & \textbf{S/H} & \textbf{S/L} & \textbf{S-pred} & \textbf{BC-pred} \\
\hline
NoXi & CPC+3DResNet & 0.64 & 0.70 & 0.65 & 0.33 \\
NoXi & TalkNet & 0.94 & 0.93 & 0.91 & 0.42 \\
NoXi & WhisperFlamingo & 0.97 & 0.91 & 0.97 & 0.45 \\
\hline
NoXi+J (Chin.+Jap.) & CPC+3DResNet & 0.50 & 0.61 & 0.51 & 0.55 \\
NoXi+J (Chin.+Jap.) & TalkNet & 0.84 & 0.82 & 0.87 & 0.62 \\
NoXi+J (Chin.+Jap.) & WhisperFlamingo & 0.85 & 0.80 & 0.93 & 0.66 \\
\hline
NoXi+J (All) & CPC+3DResNet & 0.55 & 0.59 & 0.58 & 0.39 \\
NoXi+J (All) & TalkNet & 0.85 & 0.85 & 0.89 & 0.62 \\
NoXi+J (All) & WhisperFlamingo & 0.81 & 0.86 & 0.90 & 0.44 \\
\hline
\end{tabular}%
}
\caption{F1-score results on NoXi and NoXi+J.}
\label{tab:summary_results_noxi}
\end{table}

The Haru EDR evaluation was restricted to mediation-related events because backchannel and short/long distinctions provide less direct value for the current mediation objective. The best-performing English configuration was used for each backbone. WhisperFlamingo reached 0.92 F1 for Hold, 0.85 F1 for Shift, and 0.91 F1 for S-pred, while TalkNet reached 0.84, 0.79, and 0.87, respectively. These results suggest that the model captures the floor maintenance and floor transition events that are most relevant for mediation-oriented behavior.

\begin{table}[!h]
\centering
\resizebox{\columnwidth}{!}{%
\begin{tabular}{lccc}
\hline
\textbf{Backbone} & \textbf{Hold} & \textbf{Shift} & \textbf{S-pred} \\
\hline
TalkNet & 0.83 / 0.84 & 0.78 / 0.79 & 0.87 / 0.87 \\
WhisperFlamingo & 0.92 / 0.92 & 0.83 / 0.85 & 0.91 / 0.91 \\
\hline
\end{tabular}%
}
\caption{Haru EDR results Accuracy / F1.}
\label{tab:haru_edr_mediation_results}
\end{table}

\section{Discussion and conclusions}
The proposed formulation treats turn-taking prediction as a human state-aware perception problem for social robot mediation. MM-VAP estimates the future evolution of the conversational floor from multimodal evidence. This is particularly useful for a mediator robot because the relevant action is often not to speak, but to orient, wait, avoid interruption, or prepare a balanced intervention.
The explicit mapping between predicted events and expected outputs clarifies how the model can be integrated into a future behavior layer. Active speaker detection supports gaze toward the current floor holder. Shift prediction supports anticipatory gaze toward the next speaker. Overlap detection supports gaze toward the interrupter or new floor holder. Robot turn prediction can support mediation toward the less talkative participant, while backchannel prediction should normally suppress intervention. This mapping remains a design interface and must be validated with users before being treated as a complete mediation policy.

The results obtained on NoXi, NoXi+J, and Haru EDR support the feasibility of using VA-related pretrained audio-visual encoders for MM-VAP. Nevertheless, real-time deployment remains the central open problem. The final system must demonstrate that the predictive gain is not lost through computation, buffering, synchronization, or unstable event postprocessing. Future work will therefore focus on online inference, lightweight adaptation, behavior-level validation, and integration with the complete Haru multimodal dialogue architecture. In this direction, the model is not expected to decide the complete social strategy of the robot. Instead, it should provide an estimation of the conversational floor and a small set of candidate outputs that are meaningful for mediation. Also, the more complex cases, such as deciding when the robot should verbally mediate or whether an overlap requires intervention, should remain managed by additional rules and future user studies. 

This also defines a practical discussion point for the next iteration of the work: whether predictive floor-state estimation provides enough temporal advantage to produce better human understanding and fewer inappropriate interventions than a reactive end-of-speech baseline, first in a simple 1:1 human-robot interaction and later in mediated human-human interaction.

Overall, the presented work provides a unified path between multimodal conversational perception and adaptive social robot behavior. The model is not designed to replace the robot dialogue manager, but to provide a predictive representation of the human conversational floor that can be used by a dialogue manager to decide when to observe, when to prepare, and when to mediate.


\vspace{1mm}
\noindent\textbf{Acknowledgments:} This work was partially funded by the Spanish Ministry of Science and Innovation and the State Research Agency (MCIN/AEI/10.13039/501100011033) under the projects \textbf{TIFON} [MIG-20232039 / PLEC2023-010251] (\hyperref[author:AC]{A.C.}, \hyperref[author:GP]{G.P.}), \textbf{PICRAH4.0} [PLEC2023-010353] (\hyperref[author:LM]{L.M.}) and \textbf{LIPTACON}, by CDTI Innovation through the \textit{Programa Tecnológico Espacial (PTE) 2024} via project [PTEP-20241001] (\hyperref[author:AC]{A.C.}, \hyperref[author:GP]{G.P.}). 

%%%%%%%%%%%%%%%%%%%%%%%%%%%%%%%%%%%%%%%%%%%%%%%%%%%%%%%%%%%%%%%%%%%%%%%%%%%%%%%%
%\section*{ACKNOWLEDGMENT}

%The preferred spelling of the word ÒacknowledgmentÓ in America is without an ÒeÓ after the ÒgÓ. Avoid the stilted expression, ÒOne of us (R. B. G.) thanks . . .Ó  Instead, try ÒR. B. G. thanksÓ. Put sponsor acknowledgments in the unnumbered footnote on the first page.

%%%%%%%%%%%%%%%%%%%%%%%%%%%%%%%%%%%%%%%%%%%%%%%%%%%%%%%%%%%%%%%%%%%%%%%%%%%%%%%%
\vspace{4mm}
{\setlength{\parskip}{0pt}
	\bibliographystyle{ieeetr}
	\bibliography{references}
}

\end{document}